# Explainable Artificial Intelligence and Causal Inference based ATM Fraud Detection

Yelleti Vivek[1], Vadlamani Ravi[1*], Abhay Anand Mane[2], Laveti Ramesh Naidu[2]

[1]Centre for Artificial Intelligence and Machine Learning
Institute for Development and Research in Banking Technology,
Castle Hills Road #1, Masab Tank, Hyderabad-500076, India
[2] Center for Development of Advanced Computing,
#1 Old Madras Road, Bangalore, Karnataka 560038, India
yvivek@idrbt.ac.in; vravi@idrbt.ac.in; abhaym@cdac.in; rameshl@cdac.in

**Abstract**

Customers' empathy and trust are very critical in the financial domain. Frequent occurrence of fraudulent activities affects these two factors. Hence, financial organizations and banks must take utmost care to mitigate them. Among them, ATM fraudulent transaction is a common problem faced by today's banks. There following are the critical challenges involved in fraud datasets: the dataset is highly imbalanced, the fraud pattern is changing, etc. Owing to the rarity of fraudulent activities, Fraud detection can be formulated as either a binary classification problem or One class classification (OCC). In this study, we handled these techniques on an ATM transactions dataset collected from India. In binary classification, we investigated the effectiveness of various over-sampling techniques, such as the Synthetic Minority Oversampling Technique (SMOTE) and its variants, Generative Adversarial Networks (GAN), to achieve oversampling. Further, we employed various machine learning techniques viz., Naïve Bayes (NB), Logistic Regression (LR), Support Vector Machine (SVM), Decision Tree (DT), Random Forest (RF), Gradient Boosting Tree (GBT), Multi-layer perceptron (MLP). GBT outperformed the rest of the models by achieving 0.963 AUC, and DT stands second with 0.958 AUC. DT is the winner if the complexity and interpretability aspects are considered. Among all the oversampling approaches, SMOTE and its variants were observed to perform better. In OCC, IForest attained 0.959 CR, and OCSVM secured second place with 0.947 CR. Further, we incorporated explainable artificial intelligence (XAI) and causal inference (CI) in the fraud detection framework and studied it through various analyses.

---

[*] Corresponding Author

**Table 1.** Acronyms used in the current study

| Acronym | Full form | Acronym | Full form |
|---|---|---|---|
| ATM | Automated Teller Machine | XAI | Explainable artificial intelligence |
| NB | Naïve Bayes | CI | Causal Inference |
| LR | Logistic Regression | CF | Counterfactual |
| SVM | Support Vector Machine | OCC | One class classification |
| RF | Random Forest | PSO | Particle Swarm optimization |
| DT | Decision Tree | AANN | Auto-associative neural network |
| MLP | Multi-Layer Perceptron | CRISP-DM | Cross-industry standard process for data mining |
| GBT | Gradient Boosting Tree | AANN | Auto associative Neural Network |
| SMOTE | Synthetic Minority Oversampling Technique | PSO-CARM | PSO-trained classification association rule mining |
| SMOTE-ENN | Synthetic Minority Oversampling Technique – Earliest Near Neighbour | VAE | Variational Auto Encoder |
| ADASYN | Adaptive Synthetic Sampling approach | RBF | Radial basis function |
| GAN | Generative Adversarial Network | CRISP-DM | Cross-industry standard process for data mining |
| V-GAN | Vanilla GAN | AUC | Area under the receiver operator characteristic curve |
| W-GAN | Wasserstein GAN | CV | Cross Validation |
| OCSVM | One Class Support Vector Machine | CR | Classification Rate |
| IForest | Isolation Forest | TP | True positive |
| COPOD | Copula-based Outlier Detection | FP | False positive |
| ABOD | Angle-based Outlier Detection | TN | True Negative |
| MCD | Minimum Covariance Determinant | FN | False Negative |

# 1. Introduction

Banks and Financial organizations provide various payment services to their customers, thereby making their life easier while making daily transactions. Among them, Automated Teller Machine (ATM) is one of the most popular modes of payment service. Over the past two decades, consumers across the globe have come to depend on and trust ATMs to conveniently meet their banking needs. Since then, ATMs have been facing a massive menace from fraudsters, and there is an increasing concern over these financial frauds. These fraudulent transactions inflict severe financial losses amounting to billions of dollars. These financial frauds incur a massive loss to loyal customers. Also, this will decrease the customers' empathy for the respective banks. Hence, there is a huge necessity to identify and prevent these financial frauds.

Of late, financial fraud, including credit/debit (ATM) card fraud, corporate fraud, and money laundering, has attracted the attention of all stakeholders. Across the globe, several financial companies lost billions of dollars. The Oxford English Dictionary[1] defines fraud as "*wrongful or criminal deception intended to result in financial or personal gain.*" Phua et al. (2005) describe fraud as abusing a profit organization's system without necessarily leading to direct legal consequences. Although there is no universally accepted definition of financial fraud, Wang et al. (2006) define it as "*a deliberate act that is*

---
[1] Oxford Concise English Dictionary, Tenth ed, Publisher, 1999

*contrary to law, rule, or policy with the intent to obtain an unauthorized financial benefit.*" Technology-induced frauds, such as debit (ATM)/credit card-related frauds which are particularly significant. Nowadays, none can imagine life without these financial instruments as they bring in many conveniences to customers and service providers, i.e., banks.

Owing to the importance of ATM fraud detection, in recent past years, there are various methods have been implemented. These mechanisms are not straightforward and involve multiple challenges in implementing a sophisticated ATM fraud detection methodology. One of the critical challenges these financial organisations face is that they lack end-to-end visibility into payment applications. Moreover, in today's digital world, day-to-day transactions increased enormously, resulting in millions of transactions processed monthly. Along with the above, there are a few more challenges involved. They are as follows: (i) there is a change in the fraudulent behaviour which demands a robust technique to be employed, (ii) the transactions are highly skewed/imbalanced datasets.

The challenges mentioned above can be mitigated by employing various data mining techniques and following the Cross-industry standard process for data mining (CRISP-DM) methodology[2]. There is a well-known fact that "*Data is the fuel*". Hence, to mitigate ATM fraud detection problems, one can analyze the ATM transaction history. The following scenarios can be flagged as fraudulent transactions: (i) A customer card might be stolen, and then the fraudster can initiate an abnormal sequence of transactions. (ii) Performing transactions at an unusual time. (iii) Performing transactions at an unusual location. (iv) Performing transactions with a bulk amount of money simultaneously. (v) A sequence of failed transactions initiated by the fraudster. Performing such analysis in real-time within a fraction of a second imposes an immense challenge. Hence, the following process has to be automated by employing various machine learning techniques.

Performing ATM fraud detection is indeed a classification task. Classification can be performed in various ways: (i) One class Classification: where the respective models will be trained on the negative sample data and tested on the positive class. E.g: One Class Support Vector Machine (OCSVM), Isolation Forest (IF); Whenever, there is a huge rarity of one class, one has to employ the one class classifiers. (ii) *Binary Classification*: where the classifier is trained in such a way that it can able discriminate a set of samples into two classes. As we are aware, these fraudulent datasets are highly imbalanced in nature. Hence, the imbalance in the ATM transactions data remains a scourge for almost all of the machine learning models. The same applies to other similar applications such as credit card fraud detection, churn prediction, the telecom industry, etc., where there is a hugely disproportionate between the majority class and minority class. This can be handled by employing the following techniques: (i) resampling methods, (ii) algorithmic adaptations, and (iii) cost-sensitive techniques.

In the current study, we analyzed the ATM transactions fraud detection data collected from India. There is a vast significance of this problem in real life which made us work towards this problem. In this study, we performed fraud detection using the OCC and binary classification tasks. All the acronyms used in the current study were incorporated in Table 1. We have employed several machine learning techniques to accomplish binary classification tasks: NB, LR, SVM, DT, RF, GBT, and MLP. The data imbalance is handled by employing the following techniques: SMOTE and its variants (SMOTE-ENN, SMOTE-Tomek), and ADASYN. Further, we employed the GAN variants (i.e., V-GAN and W-GAN) to achieve the oversampling. For the OCC, we employed the following techniques: OCSVM, IForest, VAE etc.; all these models and techniques were discussed in detail (refer to Section 4). Further, recently explainable artificial intelligence (XAI) occupied its prominence, especially in the sectors like healthcare, finance etc. This leverages to understand the models' behaviour, how the transactions are classified etc. Hence, we integrated XAI using SHAP into the proposed fraud detection framework. Here, we studied the features' importance, redundancy and impact on the predictions. We also employed CI and analysed the counterfactuals, which give more insights into the nature of data, which features and conditions critically entail the predictions etc., using various methods.

---

[2] CRISP DM : https://www.the-modeling-agency.com/crisp-dm.pdf

The remainder of the paper is sorted out as follows: Section 2 discusses the literature review. Section 3 presents the proposed framework for ATM transactional data using binary and OCC modelling. Section 4 presents the background theory of machine learning and the balancing techniques employed. Section 5 describes the dataset. Section 6 discusses the results of various techniques, XAI analysis, and counterfactuals. Finally, section 7 concludes the paper.

## 2. Literature Review

In this section, we will discuss the relevant studies applied to mitigate fraud detection in the area of banking and financial industries problems such as credit card fraud detection, cybersecurity-related tasks such as phishing detection, etc.,

Data mining algorithms play a quintessential role in detecting fraud post-facto (Ngai et al., 2011). These include logistic regression, neural networks, decision trees, support vector machines, etc. It can also be detected in near-real time when advanced techniques, including semi-supervised and unsupervised techniques employed to mine the data that comes in a stream (Phua et al., 2005, Edge and Sampaio, 2012). The advanced techniques typically include online classification, online clustering, and outlier detection for anomaly detection. However, the nature of the datasets would change from traditional structured-only data to unstructured and semi-structured data, which could include textual, graphical, server related data. While the former methodology is termed reactive, the latter consider much larger aspects of fraud, including transactional history. While, a reasonable amount of research is reported concerning credit card fraud detection (Ngai et al., 2011, Krishna and Ravi, 2016, Shravan Kumar and Ravi, 2015), ATM/Debit card fraud detection is seldom the researched topic. Most recently, only one research article appeared dealing with ATM transactional frauds, where a specialized language is proposed for proactive fraud management in financial data streams (Edge et al., 2012). Depending, on the usage, of an ATM card in three different ways, viz. (i) ATM, (ii) POS machine, and (iii) Internet banking. Other frauds can be perpetrated, and the way they can be detected also changes as the resulting data ranges from structured to semi-structured to unstructured. Accordingly, the challenges also vary. Wei et al. (2012) observed that the detection of card fraud is fraught with the following characteristics and challenges: (i) highly unbalanced and large, semi-structured data set; (ii) near-real-time detection; (iii) dynamic fraud behaviour; (iv) gramweak forensic evidence; and (v) diverse genuine behaviour patterns.

People often use credit cards as a mode of payment for their day-to-day transactions. Hence, fraudsters eye on it and execute several fraudulent transactions using various mechanisms. This is one of the biggest challenges the banking and financial industries face. Chan et al. (1999) are pioneers in designing a scalable solution framework for credit card fraud detection using distributed environments. The authors demonstrated the importance of the proposed solutions and the importance of consideration of scalability while designing the solution. Chen et al. (2006) proposed a binary support vector machine, a combination of SVM and genetic algorithm (GA). Here the genetic algorithm is used to select the effective combination of the support vectors. This study also incorporates the self-organizing map to estimate the input data distribution. The authors demonstrated that their approach outperformed the rest and attained a high true negative rate. Doorronsoro et al. (1997) proposed an online system for credit card fraud detection based on the neural classifier to meet real-time analytics. The authors proposed a non-linear version of Fisher's discriminant analysis to perform fraud detection. The authors proved that their approach could segregate a good proportion of fraudulent operations from normal traffic. Quah et al. (2008) proposed an online detection algorithm for real-time fraud detection where the self-organizing map is used to detect customer behaviour and fraudulent credit card transactions.

Sanchez et al. (2009) applied association rule mining to identify the anomaly behaviour pattern, which would help identify the fraudulent transaction. The authors demonstrated the effectiveness of the proposed approach on the transaction database. Srivastava et al. (2008) proposed a solution for credit card fraud detection using the Hidden Markov Model, where the customers' behaviour is used to detect credit card fraud. The authors implemented this in the following way: (i) Initially, the clustering technique is employed to cluster the set of customers with identical spending patterns. (ii) later, the hidden Markov model is executed to identify the critical patterns and sequences which capture the fraudulent transactions. Zaslavsky and Strizhak (2006) applied the self-organizing maps for credit card fraud detection.

Often, credit card transaction data poses a challenge to solve in real-time and demands the scalability concern. Syeda et al. (2002) developed a parallel granular neural network in shared memory architecture to address the above challenge. The authors applied the proposed architecture on visa card transactions to identify fraudulent transactions. The authors reported the good speed-up, and their architecture also identified the fraudulent transactions with a good amount of accuracy. Wei et al. (2012) is yet another important study that used the transaction vector to identify and store customer behaviour to handle this imbalance issue. The authors proposed a contrast miner, which effectively mines the contrast patterns and distinguishes the unlawful behaviour to mine the critical information.

Wang et al. (2006) proposed an IT-based framework to address various frauds such as computer fraud, telemarketing fraud, credit card fraud, auction fraud, etc. In their framework, the authors divided financial fraud into two categories: system attacks, and non-system attacks. The authors also developed certain security measures which might help in identifying the fraudulent activity, fraud monitoring, once the fraud is reported, how to generate the reports, etc., The authors studied the effectiveness of their framework in the case study: Taiwan financial frauds.

Due to the significant surge in the usage of online activities, several fraudsters are attacking the most often used websites by people. There are several attacks with which fraudsters can get a hold over people's online activities. Phishing is one such attack. To detect such attacks, there are several works proposed in the literature (Gaekwad, 2014). Arya et al. (2016) applied a spiking neural network inspired by neuroscience literature, which is very popular and applied to several classification problems. Such networks are applied to detect these phishing websites. Spiking neural networks outperformed the rest of the machine learning techniques. Pandey and Ravi (2012) applied to solve phishing detection by analyzing the content of emails using a document-term matrix. The authors employed various machine learning models and tested the performance of genetic programming in two ways: with and without feature selection. The authors demonstrated that DT outperformed the rest and also, owing to its white box capability, secured first in the list.

Ravisankar et al. (2011) proposed a fraud detection framework to identify the critical financial ratios obtained by analyzing the financial statements. With this, one could determine whether the business is carrying excess debt/inventory. The authors tested the proposed framework on 202 Chinese companies.

The authors demonstrated that genetic programming (GP) and probabilistic neural network (PNN) outperformed the rest and maintained similar accuracies in identifying fraudulent companies.

Often, the financial data possess missing values which hinder the performance of the models. To address this problem, Nishant et al. (2012) proposed a hybrid architecture where KNN and MLP are used in tandem. This hybrid architecture imputes the missing value and then uses the classifier to predict fraudulent activity to assess the severity of phishing attacks.

Fraquad et al. (2012) proposed a modified active learning framework where the SVM is used to extract the if-then rules and applied to solve churn prediction and insurance fraud detection. The authors proposed a three-phase approach which is described as follows: (i) SVM is used for the recursive feature elimination, (ii) synthetic data is generated by using active learning, and (iii) by using DT and NB, the rules are generated which are used to identify whether the fraud has happened or not. The data imbalance in the fraud detection datasets is handled by an under-sampling technique proposed by Vasu and Ravi (2010) in various banking and finance application. The authors proposed an under-sampling method where the K-Reverse nearest neighbourhood to remove the outliers from the majority class. Then, the classifiers are used to predict fraudulent transactions. Sundarkumar and Ravi (2013a) analyzed the dataset available in the CSMINING group. The authors invoked mutual information as the feature selection criteria are applied after extracting the features from the API calls. The dataset is imbalanced in nature. Hence the authors applied oversampling techniques and further applied various machine learning techniques.

Customers often worry about whether they are doing the transaction at the rightful website. Hence, it is essential to identify whether the website is the prey of any cyber attack. Dhanalakshmi et al. (2011) proposed a framework to identify phishing websites. Their framework analyses the server's IP address, domain, and internet domains to identify website legitimacy. Lakshmi and Vijaya (2011) proposed a supervised learning mechanism to perform phishing detection. The authors demonstrated the effectiveness of their approach by extracting the features from the 200 website URLs.

Tayal and Ravi (2016) proposed PSO-trained classification association rule mining (PSOCARM). Here, using the generated rules, one can indicate whether the phishing is detected or not. The authors proved that their approach outperformed the state-of-the-art models. Further, Tayal and Ravi (2015) also designed a fuzzy rule-based association rule mining by formulating the above as a combinatorial optimization problem. This fuzzy-based rule miner is trained by using PSO and applied to a transactional database to extract the rule, which will help to detect phishing attacks. Pradeep et al. (2014) also proposed a rule-based classifier based on firefly and threshold-accepting algorithms and applied it to classify fraudulent financial statements.

Detecting malware constitutes several challenges and demands a strong sophisticated defense mechanism. In general, malware attackers use application program interface calls (API) to steal credit card numbers, critical personal information, etc. Hence, Sundarkumar et al. (2015) used this API call information and applied topic modelling techniques to extract the information. This is followed by various ML models for malware detection. Their results prove that DT and SVM outperformed the rest, while DT is the winner in the interpretability aspect. Sheen et al. (2013) is yet another work where the features are extracted on the executable files and used an ensemble of classifiers to identify the malware. The authors focused on minimizing the number of classifiers. To achieve harmony search algorithm is used to identify the best combination set of classifiers that is productive in detecting the malware.

As we discussed earlier, fraud detection problems can be posed as OCC problems. The following are a few of the important OCC works which solved various fraud detection problems. Pandey and Ravi (2013b) are the pioneers in developing a hybrid architecture comprising particle swarm optimization (PSO) and auto-associative neural network (AANN) and named it PSO-AANN architecture. The authors posed phishing detection as a one class classification (OCC) problem. PSO-ANN outperformed the OCC classifiers, such as OC-SVM, and attained high sensitivity for the phishing email dataset. Kamaruddin and Ravi (2016) implemented the above hybrid architecture i.e., PSO-AANN architecture, under the Spark environment to make it suitable for big datasets. The proposed architecture outperformed the extant approaches, such as OC-SVM etc.,

The most often problem faced by fraud detection datasets is data imbalance. This rarity is very common in the context of OCC. To address this problem, Sundarkumar and Ravi (2015) proposed an undersampling method where the K-Reverse nearest neighbourhood and OCSVM are applied in tandem and remove the unnecessary samples in the majority class. The authors demonstrated the effectiveness of their approach by applying it to the automobile insurance fraud detection dataset and customer credit card churn prediction dataset.

Our current study is different in the following way:

- None of the above-discussed methods incorporated the XAI component in the fraud detection frameworks.
- None of the aforementioned techniques have studied the counterfactuals by subsuming the causal inference as the component.

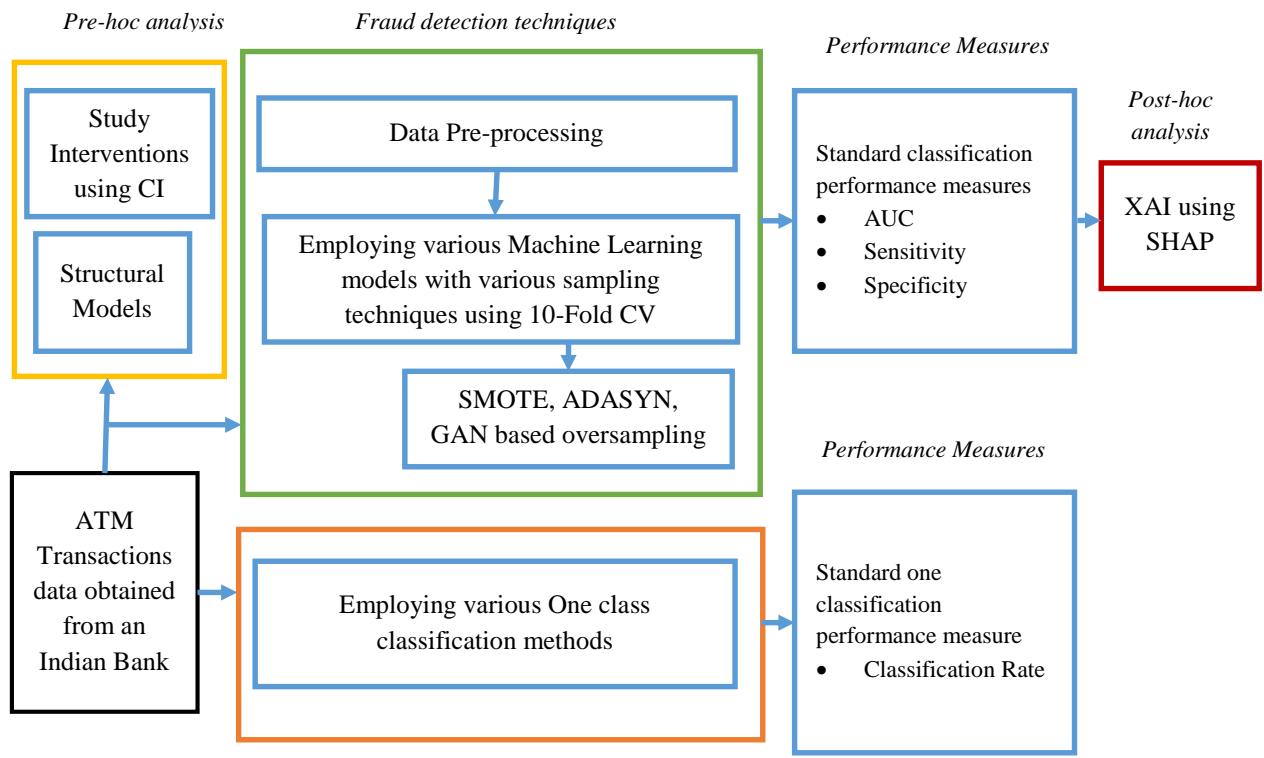

**Fig. 1. Proposed Fraud Detection Framework**

## 3. Proposed Methodology

This section discusses the proposed methodology in binary classification and OCC modelling. Firstly, we discussed concerning to the context binary classification modelling. Later, we continued the discussion on OCC modelling. The proposed fraud detection framework is depicted in Fig. 1. It includes the pre-hoc analysis where the CI is included, where the counterfactuals are studied, and then based on the chosen methodology, either the binary classification or OCC models will be evaluated. Once the model is built, using XAI to analyze feature importances, identify redundant features etc.,

### 3.1 Binary Classification Methodology

The proposed methodology is depicted in Fig. 2. We collected ATM transactional data from India. Indian Banks maintain software to monitor all ATM transactions. The dataset consists of a few days of transactions amounting to million transactions in three modes, viz., ATM transaction data, at the point of sale (POS) and in the Internet transaction. This data comprises critical information related to the type of transaction, mode of payment, customer details, transaction amount etc.; To maintain integrity and privacy of the customers we masked the feature names. Data cleansing is the process of correcting incorrect and corrupted data, removing duplicate transactions, and removing incomplete data within a dataset. The raw data extracted from the banks underwent the data cleansing to obtain a better-quality dataset. This step is crucial because '*garbage data in results in garbage analysis out*'.

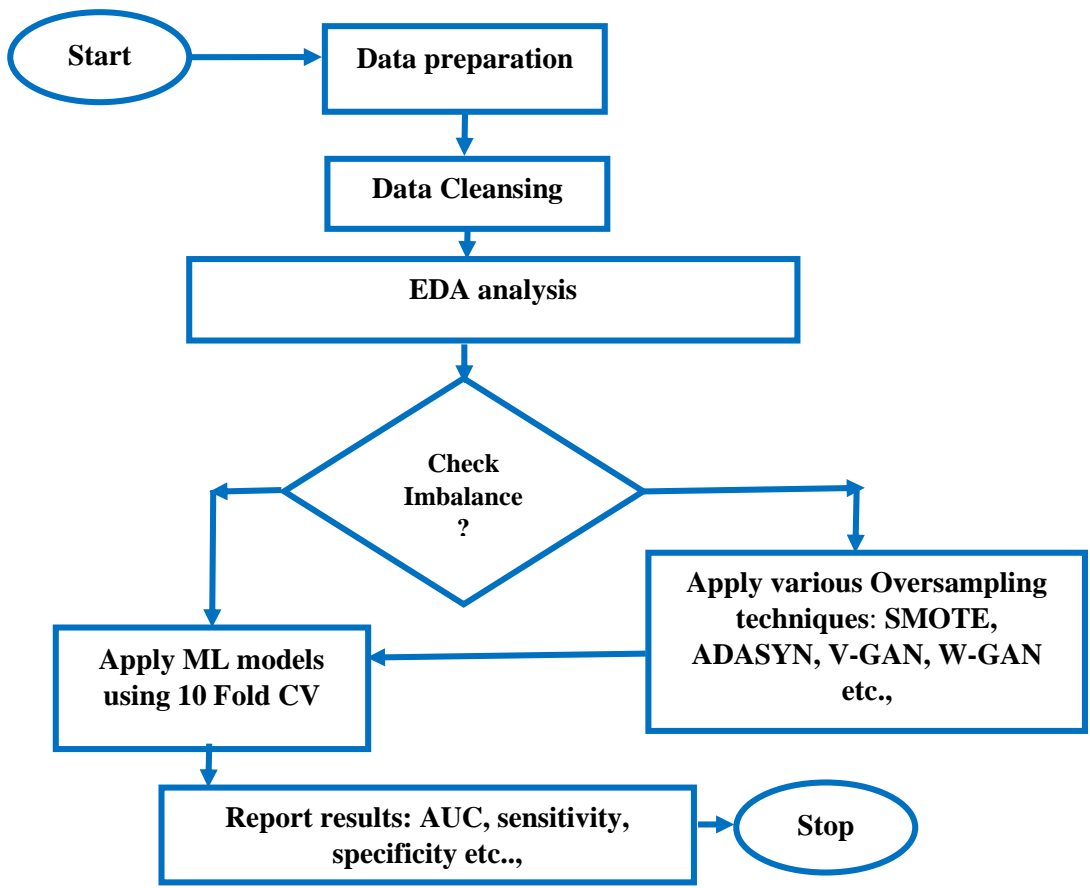

**Fig. 2. Flowchart of the proposed methodology in the context of Binary Classification**

The following steps are performed in this phase:
- All the corrupted features were fixed by following the proper mechanisms.
- All the duplicate transactions are filtered out.
- If a feature is having a percentage of null values, then these features are dropped off. We didn't perform any imputation technique here, because we observed that few of the features had almost more than 90% null values.

Once, the above step is completed, the data is in good shape to obtain insights. In this step, we analyze the data using various visual techniques. This step is crucial to discover trends, and patterns or check assumptions with the help of generated statistical summary. We observed that there is only one categorical feature and the rest of them are numerical features. The categorical feature is handled using one hot encoding technique, and the numerical features are observed to be having different ranges. Hence, we applied Normalization to the numerical features. As mentioned earlier, we renamed the features to maintain data integrity and privacy. After completing the above steps, the dataset is partitioned into training and test sets in the ratio 80:20. We followed the stratified random sampling hold-out method to perform this data partitioning. Here, stratified random sampling makes sure that the same proportion of both positive class and negative class is maintained in both training and test datasets. After performing the above steps, one has to check whether the dataset is observed to have an imbalance. If the above condition is satisfied, one has to apply the resampling techniques such as SMOTE, and ADASYN or generative resampling techniques such as V-GAN, and W-GAN. All of these sampling techniques will be discussed in detail in Section 4.2. Now, the oversampling is performed on the training dataset by concentrating on the minority class of the observed fraudulent transactions. Here, the test dataset is kept intact because it represents reality.

After completing the above step, we now build the machine learning model in this phase. In this study, we incorporated various machine learning models such as NB, LR, SVM etc., (refer to Section 4.1). The performance of the model is highly influenced by hyperparameters. Hence, while building the model, we used the grid search hyper-parameter technique. Further, we used the 10-Fold Cross Validation technique and compared the performance of the machine learning models. Once, the above process is completed, we reported the respective measures such as AUC, Sensitivity and Specificity. The complete details of these metrics will be discussed in Section 5.1.

## 3.2 OCC Methodology

Now, the discussion will be continued in the context of OCC. The proposed methodology is depicted in Fig. 3. The phases, such as data preparation, and data preprocessing, are the same, as discussed extensively in the previous section. Hence, the discussion on them was obviated.

OCC methodology works as follows: the respective OCC model is built on the negative class and tested on the positive class. Hence, unlike binary classification, the train and test split are slightly different here. The training set contains only the negative class samples, and the test set contains only the positive class samples. Once, the training and test sets are obtained the training set is normalized and then given to the modelling phase. After completing the above step, we now build the OCC model in this phase. In this study, we incorporated various machine learning models such as OCSVM, IForest etc., (refer to Section 4.3). The performance of the model is highly influenced by hyperparameters. Hence, while building the model, we used the grid search hyper-parameter technique. Once, the above process is completed, we reported the Classification Rate (CR). The complete details of these metrics will be discussed in Section 5.1.

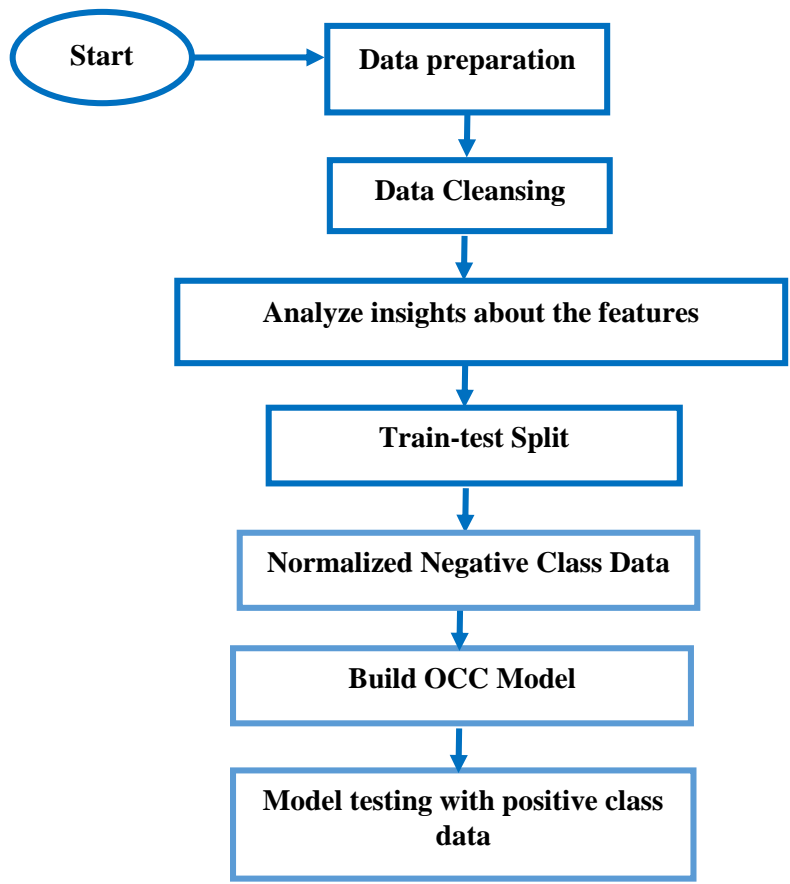

**Fig. 3. Flowchart of the proposed methodology in the context of OCC**

## 4. Overview of the techniques employed

This section briefly discusses the overall machine learning techniques employed, various balancing techniques, and one-class classification techniques in the current research.

## 4.1 Machine Learning techniques

### 4.1.1 Naïve Bayes
Naïve Bayes (NB) classifiers are the probabilistic models which are based on the Bayes theorem, are used for performing classification tasks. They are fast and easy to implement. NB works under the following assumptions: (i) all the features are independent of each other, and (ii) all the features equally contribute to the decision variable/class variable. These assumptions limit the usage of NB in real-life scenarios, where often one of these assumptions will fail thereby hindering the performance.

### 4.1.2 Logistic Regression

Logistic Regression (LR), is one of the most popular supervised classifiers for the binary classification task. This is also known as the logit model, which estimates the probability of an event occurring such as being fraudulent or legitimate, churned out or non-churn customer etc. Suppose there are n features i.e., $F = \{f_1, f_2, f_3, \ldots, f_n\}$ and y is the class variable. Then the logistic regression is mathematically represented as mentioned in Eq. (1).

$$\log(\frac{y}{1-y}) = b_o + b_1 * f_1 + b_2 * f_2 + \cdots + b_n * f_n \quad (1)$$

where $b_o$ is the bias and $b_1, \ldots, b_n$ are the coefficients of the corresponding features $f_1, \ldots, f_n$

### 4.1.3 Support Vector Machine

Support Vector Machine (SVM) proposed by Vapnik (1995), is a supervised machine learning technique used to solve various classification and regression tasks. The major objective of the SVM is to identify the optimal hyperplane which acts as a decision boundary that classifies the n-dimensional data by maximising the margin. This is nothing but structural risk minimisation. There are various kernel functions such as linear, radial basis function (RBF) etc., used in SVM. These kernel functions help in transforming the training data so the SVM could learn a non-linear decision surface.

### 4.1.4 Decision Tree

DT (Breiman et al., 1984) is one of the most powerful techniques used to solve various machine-learning tasks such as classification, and regression. DT generates a tree structure based on the training data comprising internal and leaf nodes. Each internal node serves as the classification rule, and each leaf node denotes a class label. The following are a few of the advantages of DT: handling missing values, automated feature selection, and handling both categorical and numerical features. The splitting of the root nodes is based on different criteria such as Gini, entropy etc., and DT generates a set of rules which makes it easily interpretable. Hence, DT is considered as the white box model. This model is too popular to be described in detail.

### 4.1.5 Random Forest

RF, proposed by Ho (1995), is an ensemble machine-learning technique. This is applicable for both classification and regression tasks. The training phase of RF is performed as follows: a random subset of the dataset is selected by using the bootstrap sampling technique. Upon each subset, one decision tree is built individually independent of the other. In the test phase, the test sample is passed to each decision tree which might outcome a different set of class variables. The final decision is taken by employing the majority voting principle. RF is well known for handling high-dimensional datasets effectively.

### 4.1.6 Gradient Boosting Tree

GBT is also an ensemble machine-learning technique. GBT is very well known for solving a wide variety of classification tasks such as customer churn prediction, fraud detection etc., and regression tasks. It also builds multiple decision trees. But the major difference between the RF and GBT are as follows: (i) Unlike RF, GBT builds trees one at a time, i.e. in an additive manner. Also, each new tree corrects the errors performed by the previously built trees. This technique is known as boosting. (ii) Also, the decision is taken by aggregating the so-far built tree instead of at the end as done in RF. GBT are prone to overfitting hence hyperparameter tuning has to be done very effectively.

### 4.1.7 Multi-Layer Perceptron

MLP (Rumelhart et al., 1986) is one of the most popular neural networks which maps the set of input features to a set of class variables. It is popularly known as a 'universal approximator' in solving both classification and regression problems. MLP consists of three different layers: input, hidden and output layers. The hidden layer tries to identify the non-linear patterns present in the provided dataset. The weights are estimated by using the standard backpropagation algorithm.

## 4.2 Balancing techniques

In this section, we will discuss all the balancing techniques employed in the current study.

### 4.2.1 SMOTE & its variants

**SMOTE**

SMOTE (Chawla et al., 2002), is one of the most popular and exemplary resampling methods. It duplicates the samples of the minority class and creates new synthesized samples. It first randomly selects a sample from the minority class and then uses K-nearest neighbour (KNN) to generate a new synthesized sample. This is repeated until the desired proportion of the samples were generated.

**SMOTE-ENN**

Batista et al. (2003), developed the SMOTE-ENN method which incorporates a method called Edited nearest neighbour (ENN) to remove a few of the majority class points to enrich the dataset. SMOTE-ENN is implemented in two stages where in the first stage SMOTE generates the samples, and then in the second stage, ENN removes the class points which violate its principles. ENN works by the following principle: a majority class and its K-nearest neighbour are removed from the dataset if and only if the majority class of the observations' K-nearest neighbour and the observation's class is different.

**SMOTE-Tomek**

Batista et al. (2003), developed the SMOTE-Tomek method which incorporates a method called Tomek links which acts as an undersampling technique to enrich the dataset. Tomek link is a set of two points belonging to two different classes despite being K-nearest neighbour to each other. SMOTE-Tomek is also implemented in two stages where in the first stage, SMOTE generates the samples, and in the second stage, Tomek links are removed.

### 4.2.2 ADASYN

The above-discussed SMOTE and its variants don't consider the density of the minority class while generating the synthetic samples. Hence, ADASYN (Haibo et al., 2008) is proposed by incorporating this principle thereby generating a huge number of samples in low regions and fewer / none samples in highly dense regions.

### 4.2.3 Generative Adversarial Networks

Ian Good Fellow et al. (2014) introduced GAN. Since then there are various applications GANs were used, and resampling is one of them. The main objective of the GAN is to generate data which mimics the same probability distribution as that of training data.

In our research study, we used two different variants of GAN to achieve oversampling as follows:

**V-GAN**

GAN (Ian Good Fellow et al., 2014) mainly consists of two different models, namely Discriminator and Generator respectively. Here, the generator generates the fake samples and tries to fool the discriminator, and the discriminator tries to identify the fake samples and classifies them. This makes the training of GAN, the min-max optimisation problem. The generator and the discriminator weights are optimised by freezing one at a time. This naïve version of GAN is also called Vanilla GAN.

**W-GAN**

W-GAN (Arjovsky et al., 2017) employs a different loss function known as the Wasserstein distance / Earth movers distance loss function. Wasserstein distance is used to measure the similarity between two distributions. The rest of the training procedure and remain the same for both V-GAN and W-GAN.

### 4.3 One-Class Classification techniques

Now, we will discuss the employed OCC techniques in the current study.

### 4.3.1 One class Support Vector Machine

OCSVM (Bernhard et al., 2001) is an unsupervised algorithm unlike the SVM and is used for performing class classification and outlier detection tasks. OCSVM draws the decision boundary, where the samples within the boundary are legitimate samples, and those outside the boundary are considered positive samples / fraudulent samples.

### 4.3.2 Isolation Forest

IForest (Liu et al., 2008) is an unsupervised algorithm built on decision trees and is an ensemble OCC technique. IForest also follows the bootstrap technique (like RF, refer to 3.1.5). IForest built the tree in such a way that it isolates the samples by randomly selecting a feature. Then it randomly selects the maximum and minimum values of the selected feature. Such isolated samples are identified as positive samples (or) fraudulent samples in the context of fraud detection. This process of randomly building the trees continues until the algorithm is stabilised.

### 4.3.3 Copula based outlier detection method

COPOD ( Li et al., 2020) is mainly designed to address the modelling for multivariate data distribution. This technique is inspired by copula which is used to understand the joint probabilities of the multi-variate distribution. COPOD constructs both the left and tails empirical cumulative distribution functions (CDFs). These CDFs are used to construct an empirical copula to detect the level of extremeness. The points which are identified to be at the extreme ends are considered as the positive samples.

### 4.3.4 Minimum Covariance Determinant

The MCD (Hardin and David, 2004) estimator is a highly robust estimator of the multivariate estimator and scatter. Being resistant to outlying observations makes the MCD very useful for outlier detection, and anomaly detection. MCD has also been used to develop many robust multivariate techniques, including robust principal component analysis (PCA), factor analysis and multiple regression.

### 4.3.5 Angle-based Outlier Detection

H.Kriegel et al. (2008) proposed ABOD to overcome the limitations of density-based outlier detection methods. ABOD can be used for both outlier detection and one-class classification tasks. ABOD assess the variance in the angle between the vectors of a sample to others. The farther the point, the lower the variance, which means that the probability of being this sample as the positive sample is very high. Hence, this establishes the fact that the positive samples will have relatively less variance.

### 4.3.6 Variational Auto Encoder

Knigma and Welling (2013) proposed VAE. The major objective of the VAE is to minimize the difference between a supposed distribution and the original distribution. VAE uses the KL-divergence to approximate the distribution generated and MSE to measure its closeness concerning the original dataset. VAE has mainly two components one is the encoder, and the other is the decoder. The encoder compresses the original data, and then the noise is added to this compressed data which is used by the decoder to regenerate the data which is similar to the original data. There are various applications where VAE can be used as follows: dimensionality reduction, image denoising, one-class classification, outlier detection, fraud detection etc.,

## 5 Dataset Description

In the study, we collected ATM transactional data from India. Indian Banks maintain software to monitor all ATM transactions. The dataset consists of a few days of transactions amounting to million transactions in three modes, viz., ATM transaction data, at the point of sale (POS) and in the Internet transaction. This data comprises critical information related to the type of transaction, mode of payment, customer details, transaction amount etc.; To maintain data privacy and to maintain data integrity abiding by the policies of the bank, we renamed the features. To refine the data, we followed all the data pre-processing steps for the binary classification and OCC modelling. It is observed that after refining the dataset, the total number of transactions was reduced to 7,46,724. Also, many null-valued features and unique features are dropped off, resulting in the total number of features to 10. The dataset is highly imbalanced in nature. The data proportion of non-fraud and fraud classes is 87.80:12.20. All the following experiments are performed in 10-Fold Cross Validation (CV).

## 5.1 Evaluation measures

### 5.1.1 Binary Classification Model Evaluation measures

We considered AUC for measuring the robustness of the employed classifier. In addition to AUC, we also considered sensitivity and specificity as the other measures for the binary classification.

**AUC**
It is proven to be a robust measure while handling unbalanced datasets, and is an average of specificity and sensitivity. The mathematical representation of AUC is given in Eq. (2).

$$AUC = \frac{(Sensitivity + Specificity)}{2} \qquad (2)$$

**Sensitivity**

It is also well known as True Positive Rate (TPR). It is the ratio of the positive samples that are truly predicted to be positive to all the positive samples. The mathematical notation is given in Eq. (3).

$$Sensitivity = \frac{TP}{TP + FN} \qquad (3)$$

where TP is a true positive, and FN is a false negative.

**Specificity**
It is also well known as True Negative Rate (TNR). It is the ratio of the negative samples that are truly predicted to be negative to all the negative samples. The mathematical notation is given in Eq. (4).

$$Specificity = \frac{TN}{TN + FP} \qquad (4)$$

where TN is a true negative, and FP is a false positive.

### 5.1.2 OCC Model Evaluation measures

As we know, only positive class samples are given in the test phase of OCC models. Hence, the measures mentioned above are not suitable for assessing the performance of the models. Hence, in the case of OCC, we considered classification rate (CR) as the metric to measure the quality of the OCC model.

**Classification Rate**

CR is the ratio of the number of points correctly classified as positive by the algorithm and the total number of positive samples. The mathematical representation is given in Eq. (5).

$$Classification\ rate\ (CR) = \frac{Number\ of\ correctly\ classified\ positive\ samples}{Total\ number\ of\ positive\ samples} \qquad (5)$$

Table 2: Hyperparameters for all the techniques

| Model | Hyperparameters |
|---|---|
| LR | 'regularizer': {'l1','l2','elasticnet'}<br>'optimizer': {'newton-cg','lbfgs','liblinear'} |
| SVM | 'regularizer': {'l1','l2'}<br>'loss': {'hinge','squared-hinge'} |
| DT | 'criterion': {'gini','entropy'}<br>'maxdepth': [1,2,..10] |
| RF | 'criterion': 'gini','entropy'<br>'maxdepth': [1,2,..10]<br>'estimators':[10,20,50,100,200] |
| GBT | 'loss': {'deviance','exponential'}<br>'learning_rate': [0.001,0.01,0.1]<br>'maxdepth': [1,2,..10]<br>'estimators':[10,20,50] |
| MLP | 'activation' : {'logistic','tanh','relu'}<br>'solver': {'adam','sgd'} |
| V-GAN | **Discriminator:**<br>Hidden layer 1: 128 neurons, activation function:'leaky ReLU'<br>Hidden layer 2: 64 neurons, activation function:'leaky ReLU'<br>Hidden layer 3: 32 neurons, activation function:'leaky ReLU'<br>Hidden layer 4: 8 neurons, activation function:'leaky ReLU'<br>Epochs=10,000 |
| W-GAN | **Discriminator:**<br>Hidden layer 1: 256 neurons, activation function:'leaky ReLU'<br>Hidden layer 2: 128 neurons, activation function:'leaky ReLU'<br>Hidden layer 3: 64 neurons, activation function:'leaky ReLU'<br>Hidden layer 4: 32 neurons, activation function:'leaky ReLU'<br>Epochs=10,000 |
| OCSVM | Kernel: {'linear','rbf','poly','sigmoid'} |
| IForest | n_estimators: [10,50,100,150,200,250,300]<br>max_samples: [500,1000,1500,2000] |
| ABOD | n_neighbours: [5,10,20,30,40,50] |
| VAE | **Both Encoder and Decoder**<br>Layer 1: 9 neurons; Activation Function: ReLu<br>Layer 2: 10 neurons; Activation Function: ReLu |

# 6 Results and discussion

The hyper-parameters used for the ML techniques used in the current research study are presented in Table 2. After rigorous experimentation, we presented all the hyper-parameters with their respective ranges in Table 2, which yielded the best results for all the techniques. All the experiments followed the data cleansing, and data pre-processing techniques as mentioned in Section 3. On a note, all of the balancing techniques follow a similar procedure, as mentioned in Section 4. The data proportion of non-fraud and fraud classes is 87.80:12.20. Hence, it is very imbalanced. We investigated the performance of various ML techniques by balancing with SMOTE, SMOTE-ENN, V-GAN, W-GAN etc. All the following experiments are performed in 10-Fold Cross Validation (CV).

As we are handling the imbalanced dataset, it is very well known that AUC is proven to be a robust measure, especially in case studies such as fraud detection. Hence, it is given more preference while choosing the model. Table 3 presents the AUC scores attained by ML techniques across different balancing techniques. We also presented sensitivity and specificity scores in Table 4 and Table 5, respectively.

From Table 3, one can observe that GBT attained 0.963 AUC and outperformed the rest of the models by using SMOTE-ENN as the balancing technique. The next best AUC is also attained by GBT after balancing the dataset with SMOTE-Tmoke and SMOTE, respectively. Overall, after balancing the dataset with SMOTE-ENN outperformed the rest of the balancing techniques. Interestingly, both SMOTE-Tmoke and SMOTE-ENN showed almost similar performance. Also, these two SMOTE variants performed better than the SMOTE. Hence, this owes to the fact that these two variants removed the unwanted majority of samples, thereby improving the performance of the models. After employing balancing techniques, ML techniques attained better AUC. This empirically prove that the data imbalance is a serious issue and should be dealt properly with the balancing techniques.

The second best model in terms of AUC is DT, attained 0.957, which is marginally less when compared to the best model AUC (i.e., GBT 0.963). But as a decision maker, one should not only consider the numerical best model but also the other criteria such as complexity, and interpretability. In both of these aspects, DT gets the upper hand over GBT. DT is very less complex when compared to GBT. Also, DT is a white box model. It provides 'IF-then' rules (refer to Table 6) that help to understand why the fraud might happen. As already mentioned earlier to maintain privacy we named the features to $X_1$, $X_2$, .. $X_n$ respectively. The first split happened at $X_0$. It is observed that $X_1$ played a major role in formulating the rules. By considering the above tie-breakers, now it is the responsibility of the decision-maker to choose the best model.

Table 4 shows that before balancing the dataset, all of the models attained very poor sensitivity scores, which accounts for predicting fraudulent transactions correctly. DT, RF and GBT showed 0.0 sensitivity scores, which indicate that these models are completely ineffective in predicting fraudulent transactions. After balancing the dataset, all of the models improved the sensitivity scores. This again affirms the fact observed in the context of AUC. Here also, GBT attained the best sensitivity score of 0.962, and DT stands second with a 0.954 sensitivity score. As discussed earlier, the complexity and interpretability aspect also applies here.

Considering only the numerical best to choose the best model often deceives the decision maker. Table 5 is a best example for it. There are other essential factors that need to be considered while choosing the best model. Table 5 indeed deceives the decision maker because there are several ML techniques which correctly predicted all of the non-fraudulent transactions with 100 percent specificity scores. But the same models attained very poor sensitivity scores. Especially in the case of the imbalanced dataset, DT, RF, and GBT attained a 1.0 specificity score but attained a 0.0 sensitivity score.

This is the best example model getting biased over a single class. Two more models, GBT and DT, attained 1.0 specificity scores after balancing with W-GAN. But, their corresponding sensitivity score is relatively less than their counterparts after balancing with the SMOTE-ENN technique.

Table 3: AUC scores attained by various ML techniques on 10-Fold CV

| Model | Balancing Technique | | | | | | |
|---|---|---|---|---|---|---|---|
| | Imbalanced | SMOTE | SMOTE-Tomek | SMOTE-ENN | ADASYN | V-GAN | W-GAN |
| NB | 0.746 | 0.794 | 0.793 | 0.795 | 0.676 | 0.769 | 0.742 |
| LR | 0.677 | 0.822 | 0.823 | 0.823 | 0.687 | 0.773 | 0.779 |
| SVM | 0.626 | 0.823 | 0.823 | 0.825 | 0.690 | 0.774 | 0.774 |
| DT | **0.5** | 0.955 | 0.956 | **0.958** | 0.944 | 0.907 | 0.918 |
| RF | **0.5** | 0.921 | 0.922 | 0.924 | 0.861 | 0.897 | 0.915 |
| GBT | **0.5** | 0.957 | 0.962 | **0.963** | 0.942 | 0.905 | 0.910 |
| MLP | 0.625 | 0.862 | 0.800 | 0.864 | 0.793 | 0.890 | 0.898 |

Table 4: Sensitivity scores attained by various ML techniques on 10-Fold CV

| Model | Balancing Technique | | | | | | |
|---|---|---|---|---|---|---|---|
| | Imbalanced | SMOTE | SMOTE-Tomek | SMOTE-ENN | ADASYN | V-GAN | W-GAN |
| NB | 0.763 | 0.780 | 0.779 | 0.781 | 0.542 | 0.781 | 0.726 |
| LR | 0.427 | 0.833 | 0.832 | 0.834 | 0.659 | 0.674 | 0.681 |
| SVM | 0.310 | 0.838 | 0.838 | 0.840 | 0.673 | 0.675 | 0.671 |
| DT | **0.0** | 0.950 | 0.950 | **0.953** | 0.942 | 0.828 | 0.821 |
| RF | **0.0** | 0.909 | 0.910 | 0.911 | 0.858 | 0.814 | 0.842 |
| GBT | **0.0** | 0.960 | 0.960 | **0.962** | 0.951 | 0.825 | 0.821 |
| MLP | 0.312 | 0.871 | 0.787 | 0.889 | 0.764 | 0.815 | 0.835 |

Table 5: Specificity scores attained by various ML techniques on 10-Fold CV

| Model | Balancing Technique | | | | | | |
|---|---|---|---|---|---|---|---|
| | Imbalanced | SMOTE | SMOTE-Tomek | SMOTE-ENN | ADASYN | V-GAN | W-GAN |
| NB | 0.728 | 0.808 | 0.808 | 0.809 | 0.811 | 0.758 | 0.758 |
| LR | 0.927 | 0.811 | 0.811 | 0.813 | 0.715 | 0.872 | 0.876 |
| SVM | 0.942 | 0.808 | 0.809 | 0.810 | 0.707 | 0.872 | 0.876 |
| DT | **1.0** | 0.961 | 0.962 | 0.963 | 0.945 | **0.985** | **1.0** |
| RF | **1.0** | 0.933 | 0.935 | 0.938 | 0.865 | 0.980 | 0.988 |
| GBT | **1.0** | 0.964 | 0.965 | 0.964 | 0.934 | **0.986** | **1.0** |
| MLP | 0.939 | 0.852 | 0.814 | 0.838 | 0.822 | 0.966 | 0.982 |

Table 6: Rules obtained by Decision Tree after balancing with SMOTE-ENN

| Rule No. | RULE |
|---|---|
| 1 | IF $X_0 < 0.501$ AND $0.0 < X_1 \leq 0.5$ AND $0.10 \leq X_{15} < 0.27$ THEN **Negative Class** |
| 2 | IF $X_0 < 0.501$ AND $X_1 = 0.0$ AND $X_2 \leq 0.11$ AND $X_{15} \leq 0.10$ THEN **Positive Class** |
| 3 | IF $X_0 < 0.501$ AND $0.0 < X_1 \leq 0.5$ AND $X_2 \leq 0.11$ AND $0.18 \leq X_{15} \leq 0.27$ THEN **Positive Class** |
| 4 | IF $X_0 < 0.501$ AND $X_1 = 0.0$ AND $X_{15} > 0.27$ THEN **Positive Class** |
| 5 | IF $X_0 < 0.501$ AND $X_1 = 0.0$ AND $X_2 \leq 0.11$ AND $X_{15} \leq 0.10$ THEN **Positive Class** |
| 6 | IF $X_0 < 0.501$ AND $X_1 = 0.50$ AND $X_2 \leq 0.11$ THEN **Negative Class** |
| 7 | IF $X_0 < 0.501$ AND $X_1 = 0.0$ AND $X_2 \leq 0.11$ AND $X_5 < 0.18$ THEN **Positive Class** |
| 8 | IF $X_0 < 0.501$ AND $X_1 = 0.50$ AND $X_5 \leq 0.45$ AND $X_{12} < 0.67$ THEN **Positive Class** |
| 9 | IF $X_0 < 0.501$ AND $X_1 = 0.50$ AND $X_5 \leq 0.45$ AND $X_{12} \geq 0.67$ THEN **Negative Class** |
| 10 | IF $X_0 < 0.501$ AND $0.0 < X_1 \leq 0.5$ AND $X_2 \leq 0.11$ AND $X_{15} \leq 0.27$ AND $X5 \leq 0.18$ THEN **Negative Class** |
| 11 | IF $X_0 < 0.501$ AND $0.0 < X_1 \leq 0.5$ AND $X_2 \leq 0.11$ AND $X_{15} \leq 0.27$ AND $X_5 > 0.18$ THEN **Positive Class** |
| 12 | IF $X_0 < 0.501$ AND $0.0 < X_1 \leq 0.5$ AND $X2 \leq 0.11$ AND $0.20 < X_{15} \leq 0.27$ THEN **Positive Class** |
| 13 | IF $X_0 < 0.501$ AND $0.0 < X_1 \leq 0.5$ AND $X_2 \leq 0.11$ AND $X_{15} \leq 0.20$ THEN **Negative Class** |
| 14 | IF $X_0 < 0.501$ AND $X_1 = 0.0$ AND $X_2 \leq 0.11$ AND $X_{15} \leq 0.20$ AND $X_{10} < 0.50$ THEN **Negative Class** |
| 15 | IF $X_0 < 0.501$ AND $X_1 = 0.0$ AND $X_2 \leq 0.11$ AND $X_{15} \leq 0.20$ AND $X_{10} \geq 0.50$ THEN **Positive Class** |
| 16 | IF $X_0 < 0.501$ AND $0.50 < X_1 < 0.62$ AND $X_5 = 0.0$ THEN **Negative Class** |
| 17 | IF $X_0 < 0.501$ AND $0.71 < X_1 < 0.85$ AND $X_2 < 0.24$ THEN **Negative Class** |
| 18 | IF $X_0 < 0.501$ AND $0.71 < X_1 < 0.85$ AND $X_2 < 0.24$ THEN **Positive Class** |
| 19 | IF $X_0 < 0.501$ AND $0.79 < X_1 < 0.81$ AND $X_{12} < 0.17$ THEN **Negative Class** |
| 20 | IF $X_0 < 0.501$ AND $0.79 < X_1 < 0.81$ AND $X_{12} > 0.17$ THEN **Positive Class** |
| 21 | IF $X_0 = 0.50$ AND $X_2 \leq 0.0$ AND $X_9 \leq 0.15$ THEN **Positive Class** |
| 22 | IF $X_0 = 0.50$ AND $X_2 > 0.0$ AND $X_9 > 0.15$ THEN **Negative Class** |
| 23 | IF $0.50 < X_0 \leq 0.51$ AND $X_9 \leq 0.35$ THEN **Positive Class** |
| 24 | IF $0.50 < X_0 \leq 0.51$ AND $0.25 < X_9 < 0.35$ THEN **Negative Class** |
| 25 | IF $0.51 < X_0 < 0.61$ AND $X_2 \leq 0.0$ THEN **Negative Class** |
| 26 | IF $0.62 < X_0 \leq 0.63$ AND $X_2 \leq 0.0$ THEN **Positive Class** |
| 27 | IF $0.63 < X_0 < 0.74$ AND $X_2 \leq 0.0$ AND $X_6 \leq 0.17$ THEN **Negative Class** |
| 28 | IF $0.63 < X_0 < 0.74$ AND $X_2 \leq 0.0$ AND $X_6 > 0.17$ THEN **Positive Class** |
| 29 | IF $X_0 = 0.77$ AND $0.11 \leq X_1 \leq 0.24$ THEN **Positive Class** |

Table 6: Rules obtained by Decision Tree after balancing with SMOTE-ENN (contd.)

| 30 | IF $0.78 < X_0 \leq 0.79$ AND $X_1 \leq 0.11$ AND $X_6 \leq 0.17$ THEN **Positive Class** |
|---|---|
| 31 | IF $0.78 < X_0 \leq 0.79$ AND $X_1 \leq 0.11$ AND $X_6 > 0.17$ THEN **Negative Class** |
| 32 | IF $0.82 < X_0 \leq 0.84$ AND $X_7 \leq 0.81$ AND $0.0 < X_2 \leq 0.35$ AND $X_3 \leq 0.15$ THEN **Negative Class** |
| 33 | IF $0.84 < X_0 \leq 0.86$ AND $0.0 < X_2 \leq 0.35$ AND $X_3 \leq 0.15$ THEN **Positive Class** |
| 34 | IF $X_0 > 0.84$ AND $X_2 \leq 0.08$ THEN **Negative Class** |
| 35 | IF $X_0 > 0.84$ AND $X_2 > 0.08$ THEN **Positive Class** |
| 36 | IF $(0.50 < X_0 < 0.86$ AND $X_3 \leq 0.07)$ OR $(X_5 \leq 0.11)$ THEN **Negative Class** |
| 37 | IF $(0.50 < X_0 < 0.86$ AND $X_3 \leq 0.07)$ AND $(X_5 > 0.11)$ THEN **Positive Class** |
| 38 | IF $X_0 > 0.87$ AND $X_1 \leq 0.11$ AND $X_6 \leq 0.17$ AND $X_3 \leq 0.10$ THEN **Negative Class** |
| 39 | IF $X_0 > 0.87$ AND $X_1 \leq 0.11$ AND $X_6 \leq 0.17$ AND $X_3 > 0.10$ THEN **Positive Class** |
| 40 | IF $0.87 < X_0 \leq 0.89$ AND $X_1 \leq 0.11$ AND $X_6 \leq 0.17$ AND $0.12 < X_2 < 0.20$ THEN **Positive Class** |
| 41 | IF $0.69 < X_0 \leq 0.92$ AND $X_1 \leq 0.11$ AND $X_6 \leq 0.17$ AND $X_2 > 0.12$ THEN **Positive Class** |
| 42 | IF $0.96 < X_0 < 0.99$ AND $X_1 \leq 0.11$ AND $X_6 \leq 0.17$ AND $X_3 < 0.40$ THEN **Negative Class** |
| 43 | IF $0.99 < X_0 \leq 1.0$ AND $X_1 \leq 0.11$ AND $X_6 \leq 0.17$ THEN **Positive Class** |
| 44 | IF $0.82 < X_0 \leq 0.84$ AND $X_7 \leq 0.81$ AND $0.0 < X_2 \leq 0.35$ AND $X_3 \leq 0.15$ THEN **Negative Class** |

Now, the discussion will be continued on the OCC models. All of these models were implemented by using PyOD[3] package. The results are presented in Table 7. The results show that IForest, by virtue of being an ensemble of several decision trees, outperformed the rest of the models by yielding a CR score of 0.959. The next best is attained by OCSVM which achieved a CR score of 0.947. IForest is an ensemble model and is more complex than the OCSVM. Hence, if the decision maker wants to go for less complex models, then one has to go for OCSVM, ABOD and MCD models. The choice of choosing the best algorithm is left to decision maker based on the critical factors.

Table 7: Classification Rate attained by various OCC techniques

| OCC Model | Classification Rate |
|---|---|
| OCSVM | 0.947 |
| IForest | **0.959** |
| COPOD | 0.760 |
| ABOD | 0.941 |
| MCD | 0.943 |
| VAE | 0.547 |

---

[3] https://pyod.readthedocs.io/en/latest/

## 6.1 Statistical testing of the results

Further, we conducted the t-test analysis, to statistically prove whether the superiority of the performance of the models is purely a coincidence or due to its superior nature.

The following t-test analysis is conducted on AUC obtained by the respective models during the 10-Fold CV. Here, we considered the top-performing models such as GBT, DT and RF by using the balancing technique SMOTE-ENN. Table 8 infer that p-values are less than 0.05; hence the null hypothesis is rejected, and the alternate hypothesis is accepted. We conclude that DT is more statistically significant than RF in terms of AUC, and GBT is more statistically significant than DT.

Table 8: Paired t-test results

| Model | Parameter | |
| --- | --- | --- |
| | t-statistic | p-value |
| DT vs GBT | 8.520 | $9.87 \times 10^{-08}$ |
| DT vs RF | 27.98 | $2.72 \times 10^{-16}$ |

## 6.2 Explainability using SHAP

In this section, we will discuss the eXplainability Artificial Intelligence (XAI) aspect of the trained machine learning models. Often the fraud detection frameworks ignore this aspect. But, XAI occupies its own significance as follows: understanding how the machine learning models behave, how they predict the predictions etc., This indeed increases trust and confidence.

There are various ways of studying the XAI. Among them, the most popular ones are (i) local explanations and (ii) global explanations. Among them, SHAP[4] (SHapley Additive exPlanations) is one of the most popular ones, which computes the Shapley values. Essentially, these Shapley values are the average expected marginal contribution by considering all combinations. This is performed as the post-hoc analysis. Owing to the importance of XAI, in this section will discuss the following analysis: identifying the important features, how different features are impacting the final prediction, studying the Shapley values. The following analysis is provided for the best-performing model, i.e., DT, after balancing with the SMOTE-ENN technique.

### 6.2.1 Global Feature importance

Feature importance score represents how "*important*" a feature which aids in predicting the target variable. The feature importance is depicted in Fig. 4 based on the mean SHAP values. The figure shows that $X_5$ is the more important feature, and the latter $X_1$ and $X_7$ follow it. Feature $X_{14}$, $X_{11}$, $X_{12}$ and $X_{15}$ almost attained no feature importance. Further, the complete range of all the features is also depicted in Fig. 5. Now, by observing Fig. 5, one can confirm that $X_{14}$ has showed zero impact on all the outcomes, which indicates that it has no impact in predicting an ATM transaction being fraudulent / non-fraudulent. All other features has less feature importance score which show that it is restricted to only a few number of samples. Although feature $X_1$ has more range of impact when compared to feature $X_5$, the latter has attained more dense the $X_1$. This indicates that feature $X_5$ impacts the predictions effectively than $X_1$ and this makes $X_5$ as the most important feature. It is interesting to note that $X_{10}$, $X_{11}$, and $X_{12}$ were impacted either the most or not concerning a few ATM transactions.

---

[4] https://shap.readthedocs.io/en/latest/index.html

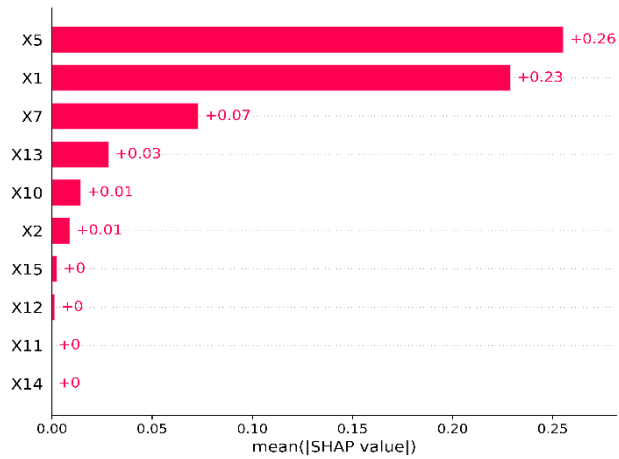

**Fig. 4. Feature importance using SHAP values**

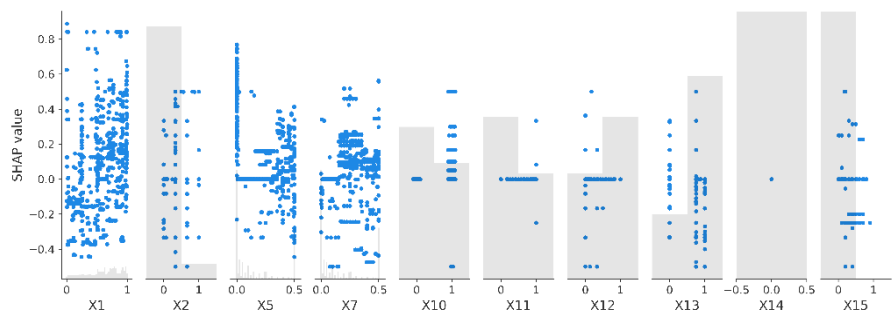

**Fig. 5. Range of SHAP values obtained by various features**

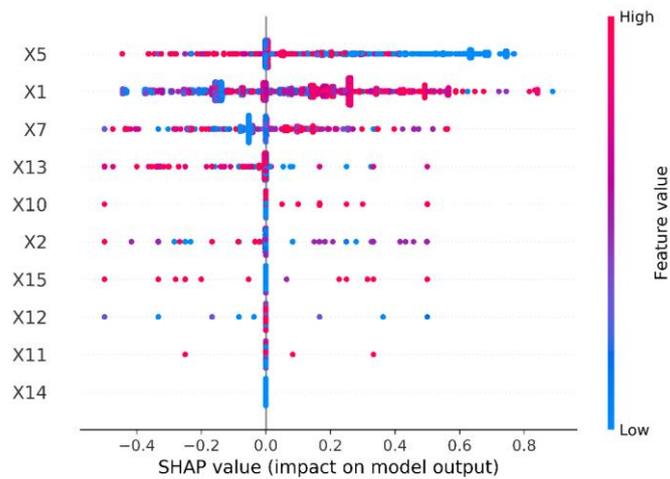

**Fig. 7. Summary plot to identify the impact of features on the predictions**

### 6.2.2 Impact of features on predictions

Along with the above analysis, we also provided the density scatter plot of SHAP values obtained by various features. These plots help in identifying the impact of each feature on the models' predictions. The density plot (refer to Fig. 7) shows that feature $X_5$ has more impact than the rest. But, the lesser dense region indicates that the impact on the predictions is less for more number of samples. Further, feature $X_1$ affected the prediction of fewer samples but at a larger amount. On the other hand, feature $X_7$ affected more predictions but in a smaller amount. Similarly, feature $X_{15}$ affected very few predictions but at a high rate. The same is observed for feature $X_{10}$.

### 6.2.3 Redundant Features

Identifying redundant features is one of the key aspects of machine learning. Hence, we conducted the below analysis by using SHAP, which considers the model loss to identify the redundant features. It considers both partial and completely redundant features. SHAP uses the hierarchical clustering method and identifies redundant features, and computes the distance, which ranges from [0,1]. The distance '0' means that they are completely irredundant features, and the distance '1' means that they are redundant features. The above analysis is depicted in Fig. 8.

In Fig. 8, there are various redundancies are observed, which are discussed below:
- $X_5$ and $X_7$ are observed to be redundant to each other. Similarly, $X_{12}$ and $X_{11}$ are observed to be redundant to each other. The same is observed among $X_1$, $X_2$, $X_{10}$, and $X_{15}$.
- In addition to the above, the subsets of features are also observed to have redundancy with other feature subsets. This is observed between $\{X_5, X_7, X_{12}, X_{11}\}$ and $\{X_{10}, X_{15}\}$. It is also observed between $\{X_5, X_7, X_{12}, X_{11}, X_{10}, X_{15}\}$ and $\{X_{13}\}$ respectively.
- It is important to note that handling redundant features is very critical. Hence, which redundancy feature combination or combinations have to be considered is left to the decision maker and the domain expert.

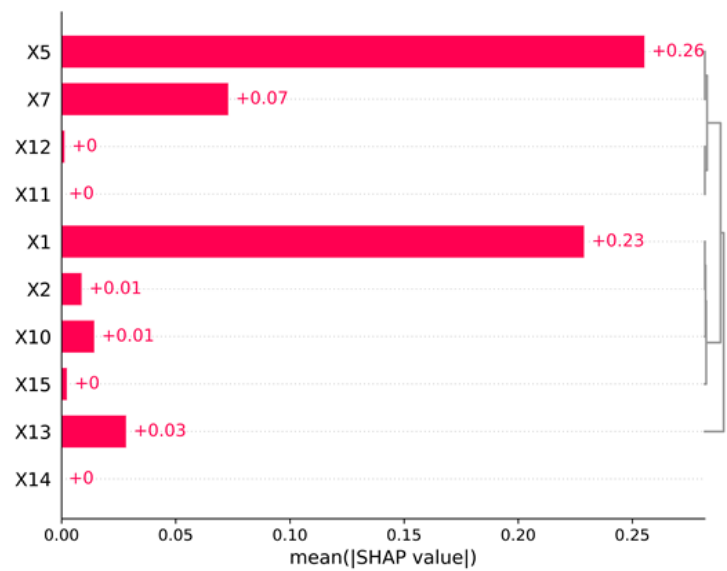

**Fig. 8. Feature redundancy by using mean SHAP values**

## 6.3 Causal inference

Causal Inference allows the researchers and practitioners to derive conclusions based on the data. Thus derived conclusions are called '*causal conclusions*', which refers to the effect of the causal variable (or treatment). Studying these interventions is critical in applications such as the medical domain, financial domains etc., where understanding the effect of varying the causal variable (or treatment) is very important. As depicted in Fig. 1, we incorporated the causal inference component as the pre-hoc analysis into the fraud detection framework. There are various methods with which these interventions can be used. Here in our study, we used various methods available in DiCE[5] to generate the counterfactuals (CFs).

DiCE provides the following methods with which we generated and studied various counterfactuals using the following methods: (i) Randomized method, (ii) KDTree and (iii) Genetic Algorithm. We conducted five different experiments to observe the counterfactuals by using the aforementioned approaches. Especially, while generating these counterfactuals, it is also vital to maintain diversity and proximity. Hence, we designed five different experiments to study various counterfactuals.

In the end, we summarized the overall observations pertaining to different methods over different experiments. A detailed analysis is provided below:

*Experiment 1: Obtaining Local importance of the features obtained on 4 different Test samples*

In this experiment, four different test dataset points were chosen to obtain the local feature importance of the feature. The same samples are used for all three approaches. It is important to note that the below mentioned local feature importance is strictly related to a single sample.

**Randomized method**

By using the Randomized method, the importance obtained is depicted in Fig. 9. The first set is related to first data point and second related to the second data point so on and so forth.

```
[{'X0': 1.0, 'X1': 1.0, 'X9': 0.9, 'X3': 0.2, 'X6': 0.2, 'X11': 0.2, 'X12': 0.1, 'X4': 0.0, 'X10': 0.0}, {'X0': 1.0, 'X1': 0.9, 'X9': 0.9, 'X4': 0.3, 'X11': 0.2, 'X10': 0.1, 'X3': 0.0, 'X6': 0.0, 'X12': 0.0}, {'X0': 1.0, 'X1': 1.0, 'X9': 1.0, 'X3': 0.3, 'X10': 0.2, 'X12': 0.2, 'X4': 0.1, 'X11': 0.1, 'X6': 0.0}, {'X0': 1.0, 'X9': 1.0, 'X1': 0.9, 'X3': 0.2, 'X4': 0.2, 'X12': 0.2, 'X6': 0.1, 'X11': 0.1, 'X10': 0.0}]
```

**Fig. 9. Randomized method: Local feature importance (screenshot taken from DiCE)**

According to this, for the first data point, $X_0$ and $X_1$ are observed to have the highest importance and $X_4$, and $X_{10}$ are observed to have no importance while deriving the target variable. The same is observed in the second point. Similarly, when considering the third point, $X_0$ and $X_9$ are observed to have the highest feature importance. The same is observed for the fourth data point as well.

**KDTree method**

Using the KDTree method, the importance obtained is depicted in Fig. 10. According to this, for the first data point, $X_0$, $X_1$, $X_3$, $X_4$, and $X_6$ are observed to have the highest importance. It is an observation that the results obtained by using the KDTree method are contradicting the results obtained using the randomized method. The same is observed for second data point.

---

[5] http://interpret.ml/DiCE/

Similarly, when considering the third point, $X_0$, $X_1$, $X_{10}$, and $X_{11}$ are observed to have the highest feature importance. However, for the fourth point, $X_0$, $X_1$, $X_3$, $X_4$, and $X_6$ are observed to have highest feature importance.

```
[{'X0': 1.0, 'X1': 1.0, 'X9': 0.9, 'X3': 0.2, 'X6': 0.2, 'X11': 0.2, 'X12': 0.1, 'X4': 0.0, 'X10': 0.0}, {'X0': 1.0, 'X1': 0.
9, 'X9': 0.9, 'X4': 0.3, 'X11': 0.2, 'X10': 0.1, 'X3': 0.0, 'X6': 0.0, 'X12': 0.0}, {'X0': 1.0, 'X1': 1.0, 'X9': 1.0, 'X3':
0.3, 'X10': 0.2, 'X12': 0.2, 'X4': 0.1, 'X11': 0.1, 'X6': 0.0}, {'X0': 1.0, 'X9': 1.0, 'X1': 0.9, 'X3': 0.2, 'X4': 0.2, 'X1
2': 0.2, 'X6': 0.1, 'X11': 0.1, 'X10': 0.0}]
```

**Fig. 10. KDtree method: Local feature importance (screenshot taken from DiCE)**

**Genetic algorithm method**

Using the Genetic algorithm method, the importance obtained is depicted in Fig. 11. For the first data point, $X_0$, $X_1$, $X_3$, $X_4$, and $X_6$ are observed to have the highest importance. Interestingly, the same set of features is identified to be important pertained to the same sample also by using the KDTree method. However, for the second point only $X_0$ feature is highly important.

```
[{'X0': 1.0, 'X1': 1.0, 'X3': 1.0, 'X4': 1.0, 'X6': 1.0, 'X9': 0.8, 'X10': 0.6, 'X11': 0.5, 'X12': 0.4}, {'X0': 1.0, 'X1': 0.9,
'X10': 0.9, 'X11': 0.9, 'X12': 0.7, 'X4': 0.5, 'X6': 0.5, 'X3': 0.3, 'X9': 0.1}, {'X0': 1.0, 'X12': 0.9, 'X3': 0.8, 'X1': 0.7,
'X4': 0.6, 'X6': 0.6, 'X9': 0.5, 'X10': 0.4, 'X11': 0.3}, {'X0': 1.0, 'X1': 1.0, 'X3': 1.0, 'X4': 1.0, 'X6': 1.0, 'X11': 0.8,
'X12': 0.7, 'X9': 0.4, 'X10': 0.4}]
```

**Fig. 11. Genetic algorithm: Local feature importance (screenshot taken from DiCE)**

Similarly, when considering the thrid point, $X_0$ is observed to have the highest feature importance. However, for the fourth point, $X_0$, $X_1$, $X_3$, $X_4$, and $X_6$ are observed to have the highest feature importance. On the contrary, the second and third data points obtained a different set of features as important features. But, again the fourth data point, the results are identical for both KDTree and genetic methods.

*Experiment 2: Counterfactuals when the desired class is opposite*

In this experiment, counterfactuals are obtained by mutating the target class as the opposite. Hence, one non-fraudulent transaction is chosen and expected to be a fraudulent transaction. Based on this, the counterfactuals were generated and studied how features are impacted by it.

**Randomized method**

The results corresponding to the randomized method are depicted in Fig. 12. The analysis is performed without keeping any restrictions on the features. Originally, the outcome of the data point is Class 0 ( non-fraudulent class). Here, we want to study how the features can be changed to make it a Class 1 (fraudulent class). The below counterfactuals show that $X_9$, $X_{10}$, $X_{11}$, and $X_{12}$ remain unchanged even though there is no restriction. The major contributed features are $X_0$, $X_1$, $X_3$, $X_4$ respectively.

It is to be noted that feature $X_0$ is a categorical feature. It is observed that in the counterfactual set, by changing the value of feature $X_0$ from 'type 4' to ' type TD' increases the chances of belonging to a positive class. Similarly, a lower value of $X_1$, $X_3$ and $X_4$ increases the probability of sample becoming a fraudulent (positive class).

```
Query instance (original outcome : 0)
   X0  X1    X3     X4     X6  X9   X10   X11  X12  label
0   4  10  10000  10000  10000   0  -15.0  15.0  0.3    0

Diverse Counterfactual set (new outcome: 1.0)
    X0  X1    X3     X4     X6  X9  X10  X11  X12  label
0   TD  10  10000  13249  10000   0  -15   15  0.3     1
1   TD  10  10000  10000  10000   0  -15   15  0.3     1
```

**Fig. 12. Randomized method: Desired class is opposite (screenshot taken from DiCE)**

**KDTree method**

The counterfactuals generated by using the KDTree method are depicted in Fig. 13. As mentioned earlier, $X_0$ is a categorical feature. The counterfactual set indicates that changing the value of feature $X_0$ from 'type 4' to 'type TD' increases the chances of belonging to a positive class. Similarly, a lower value of $X_1$, $X_3$ and $X_4$ increases the probability of sample becoming a fraudulent (positive class).

In addition, the $X_9$ value is changed from 0 to 1 also increases the probability of sample becoming fraudulent. Further, a lower $X_6$ value also increases the probability of sample becoming fraudulent. The above features are changed because they are observed to be having highest feature importance than the rest.

```
Query instance (original outcome : 0)
   X0  X1    X3     X4     X6  X9   X10   X11  X12  label
0   4  10  10000  10000  10000   0  -15.0  15.0  0.3    0

Diverse Counterfactual set (new outcome: 1.0)
    X0  X1    X3     X4     X6  X9  X10  X11  X12  label
0   TD  10  10000  13249  10000   0  -15   15  0.3     1
1   TD  10  10000  10000  10000   0  -15   15  0.3     1
```

**Fig. 13. KDtree method: Desired class is opposite (screenshot taken from DiCE)**

**Genetic algorithm method**

The counterfactuals generated by the Genetic algorithm are depicted in Fig. 14. Interestingly, the counterfactuals generated by the Genetic algorithm are close to that of the KDTree method.

By changing the value of feature $X_0$ from 'type 4' to 'type TD' increases the chances of belonging to a positive class. Similarly, a lower value of $X_1$, $X_3$ and $X_4$ increases the probability of sample becoming a fraudulent (positive class). Further, the $X_9$ value is changed from 0 to 1 also increases the probability of sample becoming fraudulent. In addition, by maintaining a lower $X_6$ value also increases the probability of sample becoming fraudulent.

The above features are changed because they are observed to be having highest feature importance than the rest.

```
   X0  X1    X3      X4      X6    X9  X10   X11  X12  label
0  4   10  10000.0  10000.0  10000.0  0  -15.0  15.0  0.3    0
```

Diverse Counterfactual set (new outcome: 1)

```
   X0  X1   X3     X4     X6   X9  X10   X11   X12  label
0  TD  0   5000.0 5000.0 5000.0  1  -15.0 15.0  0.30   1
0  TD  0   2030.0 1518.0  0.0   0  -15.0 15.0  0.35   1
```

**Fig. 13. Genetic algorithm method: Desired class is opposite (screenshot taken from DiCE)**

*Experiment 3: Counterfactuals when the desired class is opposite by keeping constraints on one random feature*

In this experiment, counterfactuals are obtained by mutating the target class as the opposite in a constrained environment. The constraints are kept on a randomly selected feature. Then, the counterfactuals were generated and studied how other features are impacted by it.

**Randomized method**

The results corresponding to the randomized method are depicted in Fig. 14. The analysis is performed by keeping a restriction over one randomly selected feature. For our analysis, we have chosen $X_3$. We kept the constraint as $X_3$ lies between the range of [0, 1000]. Here, also the desired outcome is the opposite class. As the query passed is a negative class the desired outcome is positive. The same feature is chosen for other methods as well.

The counterfactuals (refer to Fig. 14), show that $X_0$ plays a major role. When the $X_0$ category is 'type TD', the probability of a sample becoming fraudulent increases. Further, when the $X_1$ value is close to 0, the probability of sample becoming fraudulent is very high. A lower $X_{12}$ value also influences tends sample to become fraudulent.

Query instance (original outcome : 0)
```
   X0  X1   X3    X4    X6  X9  X10   X11  X12  label
0  4   10 10000 10000 10000  0  -15.0 15.0  0.3    0
```

Diverse Counterfactual set (new outcome: 1.0)
```
   X0  X1   X3    X4 X6 X9 X10  X11  X12  label
0  TD  10   -     -  -   0  -   0.0  0.3    1
1  TD  0    -     -  -   0  -    -   0.3    1
2  TD  -    -     -  -   0  -    -   0.3    1
3  TD  10  7510.0 -  -   0  -    -   0.3    1
4  TD  10   -     -  -   0  -    -   0.14   1
```

**Fig. 14. Randomized method: Desired class is opposite with constraints over a feature (screenshot taken from DiCE)**

**KDTree method**

The results corresponding to KDTree is depicted in Fig. 15. Here, also the desired outcome is opposite class. As the query passed is negative class the desired outcome is positive.

|    | X0 | X1 | X3 | X4 | X6 | X9 | X10 | X11 | X12 | label |
|----|----|----|----|----|----|----|----|----|----|----|
| 0 | 4 | 10 | 10000 | 10000 | 10000 | 0 | -15.0 | 15.0 | 0.3 | 0 |

Diverse Counterfactual set (new outcome: 1)

|    | X0 | X1 | X3 | X4 | X6 | X9 | X10 | X11 | X12 | label |
|----|----|----|----|----|----|----|----|----|----|----|
| 627987 | TD | 0 | 5000.0 | 5000.0 | 5000.0 | 1 | - | - | 0.3 | 1 |
| 518613 | TD | 0 | 5000.0 | 5000.0 | 5000.0 | 1 | - | - | 0.35 | 1 |
| 643226 | TD | 0 | 5000.0 | 0.0 | 0.0 | - | 0.0 | 0.0 | 0.0 | 1 |
| 611676 | TD | 0 | 5000.0 | 0.0 | 0.0 | 1 | 0.0 | 0.0 | 0.0 | 1 |
| 165898 | TD | 0 | 3000.0 | 0.0 | 0.0 | 1 | 0.0 | 0.0 | 0.0 | 1 |

**Fig. 15. KDTree method: Desired class is opposite with constraints over a feature (screenshot taken from DiCE)**

The counterfactuals (refer to Fig. 15), show that $X_0$ plays a major role (derived this by using feature importance experiment). When the $X_0$ category is 'type TD', the probability of sample becoming a fraudulent increases. Also, when the $X_1$ value is close to 0, the probability of sample becoming fraudulent. A lower $X_{12}$ value also influences of sample becoming fraudulent. Further, when $X_4$ is closer to zero and $X_9$ tends to 1, then in the majority of the counterfactuals has shown that the transaction becomes a fraudulent one.

**Genetic algorithm method**

The results corresponding to the Genetic algorithm are depicted in Fig. 9. The below counterfactuals (refer to Fig. 9), shows that $X_0$ plays a major role (derived this by using feature importance). When the $X_0$ category is TD, the probability of sample becoming a fraudulent increases. Also, when the $X_1$ value is close to 0, the probability of sample becoming a fraudulent is more. A lower $X_{12}$ value also influences of becoming a fraudulent sample. Further, if $X_9$ tends to 1, and then the majority of the counterfactuals have shown that the transaction becomes a fraudulent one. Interestingly, the counterfactuals obtained by the Genetic algorithm are the same as the KDTree method.

|    | X0 | X1 | X3 | X4 | X6 | X9 | X10 | X11 | X12 | label |
|----|----|----|----|----|----|----|----|----|----|----|
| 0 | 4 | 10 | 10000 | 10000 | 10000 | 0 | -15.0 | 15.0 | 0.3 | 0 |

Diverse Counterfactual set (new outcome: 1)

|    | X0 | X1 | X3 | X4 | X6 | X9 | X10 | X11 | X12 | label |
|----|----|----|----|----|----|----|----|----|----|----|
| 627987 | TD | 0 | 5000.0 | 5000.0 | 5000.0 | 1 | - | - | 0.3 | 1 |
| 518613 | TD | 0 | 5000.0 | 5000.0 | 5000.0 | 1 | - | - | 0.35 | 1 |
| 643226 | TD | 0 | 5000.0 | 0.0 | 0.0 | - | 0.0 | 0.0 | 0.0 | 1 |
| 611676 | TD | 0 | 5000.0 | 0.0 | 0.0 | 1 | 0.0 | 0.0 | 0.0 | 1 |
| 165898 | TD | 0 | 3000.0 | 0.0 | 0.0 | 1 | 0.0 | 0.0 | 0.0 | 1 |

**Fig. 16. Genetic algorithm: Desired class is opposite with constraints over a feature (screenshot taken from DiCE)**

*Experiment 4: Generating counterfactuals by maintaining proximity and diversity*

**Randomized method**

```
Query instance (original outcome : 0)
   X0  X1    X3     X4     X6  X9  X10   X11  X12  label
0   4  10  10000  10000  10000   0 -15.0  15.0  0.3    0

Diverse Counterfactual set (new outcome: 1.0)
   X0  X1    X3     X4  X6  X9  X10  X11  X12  label
0  TD  10  4394.0      -   -   0    -    -  0.3      1
1  TD  10       -  5041.0  -   0    -    -  0.3      1
2  TD  10 11260.0      -   -   0    -    -  0.3      1
3  TD  10       -      -   -   0 -11.0   -  0.3      1
4  TD  10       -      -   -   0  -6.6   -  0.3      1
5  TD  10       -  2669.0  -   0    -    -  0.3      1
6  TD   2       -      -   -   0    -    -  0.3      1
7  TD  10   243.0      -   -   0    -    -  0.3      1
8  TD  20       -      -   -   0    -    -  0.3      1
9  TD  10  3633.0      -   -   0    -    -  0.3      1
```

**Fig. 17. Randomized method: generating CFs by maintaining proximity and diversity (screenshot taken from DiCE)**

The results corresponding to random are depicted in Fig. 17. The analysis is performed by giving proximity weightage as 1.0 and diversity weightage as 3.0. Here, also the desired outcome is the opposite class. The same parameters are maintained for other methods as well.

The below counterfactuals (refer to Fig. 17), show that:
- $X_0$ is changed from 4 to TD type.
- $X_1$ value is decreased. Hence, a lower X1 increases the risk of transaction becoming fraudulent.
- In 40% of the generated counterfactuals, the $X_3$ value also lowered. Hence lowering $X_3$ may become the transaction as fraudulent. But 60% deny it hence the effect is nominal not as devastating as $X_0$, and $X_1$.
- Higher $X_{10}$ may increase the probability of transaction becoming a fraudulent.

**KDTree method**

The results corresponding to KDTree are depicted in Fig. 11. As mentioned earlier, proximity weightage and diversity weightage are maintained as 1.0 and 3.0 respectively. The below counterfactuals (refer to Fig. 18), show that:
- $X_0$ is changed from 'type 4' to 'type TD' then it becomes a fraudulent transaction.
- $X_1$ value is zero then the risk of transaction becoming fraudulent is very high.
- The above $X_1$ is observed for $X_4$ as well.
- If $X_{11}$ and $X_{12}$ values were lower then the probability of transaction becoming a fraudulent is very high.
- As per the importance the first two inferences w.r.t the $X_0$ and $X_1$ are sufficient, and hence those two features will play an important role.

|  | X0 | X1 | X3 | X4 | X6 | X9 | X10 | X11 | X12 | label |
|---|---|---|---|---|---|---|---|---|---|---|
| 0 | 4 | 10 | 10000 | 10000 | 10000 | 0 | -15.0 | 15.0 | 0.3 | 0 |

Diverse Counterfactual set (new outcome: 1)

|  | X0 | X1 | X3 | X4 | X6 | X9 | X10 | X11 | X12 | label |
|---|---|---|---|---|---|---|---|---|---|---|
| 627987 | TD | 0 | 5000.0 | 5000.0 | 5000.0 | 1 | - | - | 0.3 | 1 |
| 518613 | TD | 0 | 5000.0 | 5000.0 | 5000.0 | 1 | - | - | 0.35 | 1 |
| 643226 | TD | 0 | 5000.0 | 0.0 | 0.0 | - | 0.0 | 0.0 | 0.0 | 1 |
| 611676 | TD | 0 | 5000.0 | 0.0 | 0.0 | 1 | 0.0 | 0.0 | 0.0 | 1 |
| 165898 | TD | 0 | 3000.0 | 0.0 | 0.0 | 1 | 0.0 | 0.0 | 0.0 | 1 |
| 524548 | TD | 0 | 2500.0 | 0.0 | 0.0 | 1 | 0.0 | 0.0 | 0.0 | 1 |
| 225800 | TD | 0 | 1000.0 | 0.0 | 0.0 | 1 | 0.0 | 0.0 | 0.0 | 1 |
| 262430 | TD | 0 | 500.0 | 0.0 | 0.0 | 1 | 0.0 | 0.0 | 0.0 | 1 |
| 290706 | TD | 0 | 70.0 | 0.0 | 0.0 | 1 | 0.0 | 0.0 | 0.0 | 1 |

**Fig. 18. KDtree method: generating CFs by maintaining proximity and diversity (screenshot taken from DiCE)**

**Genetic algorithm method**

The results corresponding to the Genetic algorithm are depicted in Fig. 19. As mentioned earlier, proximity weightage and diversity weightage are maintained as 1.0 and 3.0 respectively. The below counterfactuals (refer to Fig. 19), show that:

- $X_0$ is changed from 'type 4' to 'type TD' type then it becomes a fraudulent transaction.
- $X_1$ value is close to zero or zero then the risk of transaction becoming fraudulent is very high.
- The above $X_1$ is observed for $X_4$ as well.
- If the $X_9$ value is mutated then the probability of transaction becoming fraudulent also increases. This is a vital observation and didn't obtain in the counterfactuals obtained by using the randomized method / KDTree method.
- If $X_{11}$ were lower then the probability of transaction becoming fraudulent is very high.

Diverse Counterfactual set (new outcome: 1.0)

|  | X0 | X1 | X3 | X4 | X6 | X9 | X10 | X11 | X12 | label |
|---|---|---|---|---|---|---|---|---|---|---|
| 0 | TD | 10 | - | - | - | 0 | - | - | 0.3 | 1 |
| 1 | TD | 10 | - | 6012.0 | - | 0 | - | - | 0.3 | 1 |
| 2 | TD | 10 | 2.0 | - | - | 0 | - | - | 0.3 | 1 |
| 3 | TD | 10 | - | - | - | - | - | - | 0.3 | 1 |
| 4 | TD | 0 | - | - | - | 0 | - | - | 0.3 | 1 |
| 5 | TD | 10 | - | - | - | 0 | - | - | 0.11 | 1 |
| 6 | TD | 10 | - | - | - | 0 | - | - | 0.41 | 1 |
| 7 | TD | 10 | - | - | 7627.0 | 0 | - | - | 0.3 | 1 |
| 8 | TD | 10 | - | - | - | 1 | - | - | 0.3 | 1 |
| 9 | TD | 10 | - | - | - | 0 | -13.9 | - | 0.3 | 1 |

**Fig. 19. Genetic algorithm: generating CFs by maintaining proximity and diversity (screenshot taken from DiCE)**

*Experiment 5: Generating counterfactuals by maintaining proximity and diversity while constraining a randomly selected feature.*

The analysis is performed by giving proximity weightage as 1.0 and diversity weightage as 3.0. The feature $X_3$ is selected, and the constraints kept over $X_3$ are [1,5]. Here, also the desired outcome is the opposite class. The same settings are used for other methods also.

**Randomized method**

The results corresponding to random are depicted in Fig. 20. The observations of counterfactuals are as follows:
- $X_0$ is changed from 'type 4' to 'type TD' type then it becomes a fraudulent transaction.
- $X_1$ value is close to zero or zero then the risk of transaction becoming fraudulent is very high.
- As a surprise, lower $X_{12}$ values are observed to have fraudulent classes. But now even the higher value of $X_{12}$ also gave the fraudulent class.

Diverse Counterfactual set (new outcome: 1.0)

|   | X0 | X1 | X3  | X4     | X6     | X9 | X10   | X11 | X12  | label |
|---|----|----|-----|--------|--------|----|-------|-----|------|-------|
| 0 | TD | 10 | -   | -      | -      | 0  | -     | -   | 0.3  | 1     |
| 1 | TD | 10 | -   | 6012.0 | -      | 0  | -     | -   | 0.3  | 1     |
| 2 | TD | 10 | 2.0 | -      | -      | 0  | -     | -   | 0.3  | 1     |
| 3 | TD | 10 | -   | -      | -      | -  | -     | -   | 0.3  | 1     |
| 4 | TD | 0  | -   | -      | -      | 0  | -     | -   | 0.3  | 1     |
| 5 | TD | 10 | -   | -      | -      | 0  | -     | -   | 0.11 | 1     |
| 6 | TD | 10 | -   | -      | -      | 0  | -     | -   | 0.41 | 1     |
| 7 | TD | 10 | -   | -      | 7627.0 | 0  | -     | -   | 0.3  | 1     |
| 8 | TD | 10 | -   | -      | -      | 1  | -     | -   | 0.3  | 1     |
| 9 | TD | 10 | -   | -      | -      | 0  | -13.9 | -   | 0.3  | 1     |

**Fig. 20. Randomized method: generating CFs by maintaining proximity and diversity by constraining a feature (screenshot taken from DiCE)**

**KDTree method**

Interestingly, by following the above settings resulted in the generation of zero counterfactuals. Hence, this establishes the fact that if a method cannot find the counterfactuals, then it will not generate any counterfactuals.

**Genetic algorithm method**

The below counterfactuals (refer to Fig. 21), show that:
- $X_0$ is changed from ' type 4' to 'type TD' then it becomes a fraudulent transaction.
- $X_1$ value is either zero to close to zero then the risk of transaction becoming fraudulent is very high.
- $X_9$ is mutated and chosen to be 1. Then the probability of transaction becoming fraudulent is very high.

- When $X_4$ and $X_6$ are maintained to having identical values then the risk of a transaction becoming fraudulent is also very high.

| | X0 | X1 | X3 | X4 | X6 | X9 | X10 | X11 | X12 | label |
|---|---|---|---|---|---|---|---|---|---|---|
| 0 | TD | 0 | 14138.0 | 2776.0 | 13905.0 | 1 | -15.0 | 15.0 | 0.30 | 1 |
| 0 | TD | 0 | 5000.0 | 5000.0 | 5000.0 | 1 | -15.0 | 15.0 | 0.30 | 1 |
| 0 | TD | 10 | 0.0 | 5232.0 | 9633.0 | 1 | -15.0 | 7.0 | 0.30 | 1 |
| 0 | TD | 0 | 5000.0 | 5000.0 | 19804.0 | 1 | -15.0 | 4.0 | 0.30 | 1 |
| 0 | TD | 3 | 14138.0 | 0.0 | 9058.0 | 0 | -3.0 | 0.0 | 0.30 | 1 |
| 0 | TD | 0 | 5000.0 | 5000.0 | 5000.0 | 1 | -15.0 | 15.0 | 0.35 | 1 |
| 0 | TD | 0 | 5000.0 | 2776.0 | 5000.0 | 1 | -1.0 | 7.0 | 0.35 | 1 |
| 0 | TD | 3 | 0.0 | 2776.0 | 5000.0 | 1 | -1.0 | 0.0 | 0.35 | 1 |
| 0 | TD | 0 | 0.0 | 4337.0 | 0.0 | 0 | -15.0 | 15.0 | 0.00 | 1 |
| 0 | TD | 20 | 9844.0 | 0.0 | 15202.0 | 0 | -15.0 | 15.0 | 0.00 | 1 |

**Fig. 21. Genetic algorithm: generating CFs by maintaining proximity and diversity by constraining a feature (screenshot taken from DiCE)**

In summary, we had observed the following important observations:

1. Randomized method is pretty straightforward and often takes lesser time to generate the counterfactuals. But as we know that these counterfactuals are randomly generated which is a disadvantage with this method. In the systems where the arbitrary counterfactuals this method has to be employed owing to its simplicity.
2. On the other hand, counterfactuals generated by KDTree and Genetic algorithm methods are more complex. In fact, the latter is more complex than the former. To preserve the domain knowledge and whenever one want to check the interventions caused in a specific interval, these two methods will come handy.

## 7. Conclusions and Limitations

In this study, we investigated the identification of fraudulent transactions in the ATM transactions dataset collected from India. We posed this problem as both a binary classification dataset and OCC. In binary classification, we over-sampled the minority samples by using various techniques viz., Synthetic Minority Oversampling Technique (SMOTE) and its variants, Generative Adversarial Networks (GAN) also to achieve oversampling. Further, we employed various machine learning techniques where GBT outperformed the rest of the models by achieving 0.963 AUC, and DT stands second with 0.958 AUC. DT is the winner if the complexity and interpretability aspects are preferred to be tie-breakers. Among all of the oversampling approaches, SMOTE and its variants outperformed the rest. In OCC, IForest attained 0.959 CR and OCSVM secured second place with 0.947 CR. Further, we incorporated explainable artificial intelligence (XAI) and causal inference (CI) in the fraud detection framework and studied it through various analyses.

In future, the same problem can be solved by providing scalable solutions to meet today's big data world. Further, the current study is limited to solving under a static environment. Hence, it can be further posed as a real-time analytics problem and can be solved by using streaming analytics. Recently, Kate et al. proposed Chaotic Fin-GAN for financial domain problems. Hence, incorporating Chaotic Fin-GAN is also a potential future work.